\documentclass[acmsmall]{acmart}

\AtBeginDocument{%
  \providecommand\BibTeX{{%
    \normalfont B\kern-0.5em{\scshape i\kern-0.25em b}\kern-0.8em\TeX}}}

\usepackage{multirow}
\usepackage{array}
\usepackage{setspace}
\usepackage{url}
\usepackage{graphicx}
\usepackage{epstopdf}
\usepackage{bm}
\usepackage{makecell}
\newtheorem{myDef}{Definition}
\begin{document}

\title{Enabling Harmonious Human-Machine Interaction with Visual-Context Augmented Dialogue System: A Review}


\author{Hao Wang}
\email{wanghao456@mail.nwpu.edu.cn}
\affiliation{%
  \institution{Northwestern Polytechnical University}
  \city{Xi'an}
  \country{P.R.China}}

\author{Bin Guo}
\email{guob@nwpu.edu.cn (Corresponding-author)}
\authornotemark[1]
\affiliation{%
  \institution{Northwestern Polytechnical University}
  \city{Xi'an}
  \country{P.R.China}
}

\author{Yating Zeng}
\email{1142113183@qq.com}
\affiliation{%
  \institution{Northwestern Polytechnical University}
  \city{Xi'an}
  \country{P.R.China}}

\author{Yasan Ding}
\email{dingyasan@163.com}
\affiliation{%
  \institution{Northwestern Polytechnical University}
  \city{Xi'an}
  \country{P.R.China}}

\author{Chen Qiu}
\email{qiuchen@nwpu.edu.cn}
\affiliation{%
  \institution{Northwestern Polytechnical University}
  \city{Xi'an}
  \country{P.R.China}}

\author{Ying Zhang}
\email{izhangying@nwpu.edu.cn}
\affiliation{%
 \institution{Northwestern Polytechnical University}
 \city{Xi'an}
 \country{P.R.China}}

\author{Lina Yao}
\email{lina.yao@unsw.edu.au}
\affiliation{%
  \institution{The University of New South Wales}
  \city{Sydney}
  \country{Australia}
}

\author{Zhiwen Yu}
\email{zhiwenyu@nwpu.edu.cn}
\affiliation{%
  \institution{Northwestern Polytechnical University}
  \city{Xi'an}
  \country{P.R.China}}

%\renewcommand{\shortauthors}{Trovato and Tobin, \textit{et al.} }
\renewcommand{\shortauthors}{H.Wang, et al.}
\begin{abstract}

The intelligent dialogue system, aiming at communicating with humans harmoniously with natural language, is brilliant for promoting the advancement of human-machine interaction in the era of artificial intelligence. With the gradually complex human-computer interaction requirements (e.g., multimodal inputs, time sensitivity), it is difficult for traditional text-based dialogue system to meet the demands for more vivid and convenient interaction. Consequently, Visual-Context Augmented Dialogue System (VAD), which has the potential to communicate with humans by perceiving and understanding multimodal information (i.e., visual context in images or videos, textual dialogue history), has become a predominant research paradigm. Benefiting from the consistency and complementarity between visual and textual context, VAD possesses the potential to generate engaging and context-aware responses. For depicting the development of VAD, we first characterize the concepts and unique features of VAD, and then present its generic system architecture to illustrate the system workflow. Subsequently, several research challenges and representative works are detailed investigated, followed by the summary of authoritative benchmarks. We conclude this paper by putting forward some open issues and promising research trends for VAD, e.g., the cognitive mechanisms of human-machine dialogue under cross-modal dialogue context, and knowledge-enhanced cross-modal semantic interaction.

\end{abstract}



\setcopyright{acmcopyright}
\acmJournal{CSUR}
\acmYear{2019} 
\acmVolume{1} 
\acmNumber{1} 
\acmArticle{1} 
\acmMonth{1} 
\acmPrice{15.00}


\maketitle

\section{Introduction}
%The development of deep learning techniques has contributed to the unprecedented advancement of many fields, such as Computer Vision (CV), Natural Language Processing (NLP) and Speech Recognition. Researches of each filed usually focus on a specific modal. Processing and understanding images is the main target of CV, leading to the outstanding performance breakthroughs in image classification \cite{wang2017residual}, object detection \cite{yu2016unitbox}, face recognition \cite{schroff2015facenet}, and other image-based tasks \cite{voulodimos2018deep}. Moreover, NLP is dedicated to allowing machines to process and understand natural language, thus enabling communication between humans and machines, such as sentiment analysis \cite{xu2019bert}, question answering systems \cite{lee2019latent} and dialogue systems \cite{zhang2020dialogpt}. The breakthroughs in these domains have brought us numerous convenient services. For example, based on face information for identity authentication, face recognition has widespread applications in intelligent security \cite{kumar2019intelligent}, face payment \cite{nasution2020face} and other scenarios. Dialogue systems can perform specific tasks for users or communicate with us using natural language as humans, and have already been widely deployed in real world, such as Microsoft XiaoIce\footnote{\url{http://www.msxiaobing.com/}}, Cortana\footnote{\url{http://www.msxiaona.cn/}} and Apple Siri\footnote{\url{https://www.apple.com/siri/}}.

%直接从对话系统开始说起，不说深度学习了
The dialogue system, also known as chatbot, is a representative approach of human-machine interaction, which can complete specific tasks (e.g., movie tickets booking, smart devices controlling) and chitchat with humans in natural language. Building intelligent dialogue systems that can naturally and engagingly converse with humans has been a long-standing goal of artificial intelligence (AI) \cite{gao2019neural, huang2020challenges}. The cumulative high-quality conversational data \cite{lowe2015ubuntu, li2017dailydialog, wang2021naturalconv} and the emergence of large-scale natural language processing (NLP) models \cite{devlin2019bert, radford2018improving} have facilitated to the rapid development of dialogue systems. Existing dialogue systems are empowered with the potential to generate coherent, logical, and even personalized and informative conversational responses \cite{mazare2018training, young2018augmenting, zhang2020dialogpt}, which provide diverse and convenient interactive services to users. For example, popular intelligent assistants (e.g., Apple Siri\footnote{\url{https://www.apple.com/siri/}}, Microsoft Cortana\footnote{\url{http://www.msxiaona.cn/}}) can assist us to operate smartphones and computers by voice for completely hands free, and open-domain chatbots (e.g., Microsoft XiaoIce\footnote{\url{http://www.xiaoice.com/}}) can communicate with humans about any topics and grow into intimate virtual companions for users.

%从文本对话系统引入到多模态对话系统
We humans will acquire the knowledge information from cross-media data, such as text, image, video, audio. And human-human interaction is naturally multimodal, where we understand context information embodied in textual language, audio, and visual scenes to express emotion, mood, attitude, and attention about the surroundings. Therefore, perceiving and understanding multimodal context information is essential for natural and harmonious human-machine interaction systems \cite{jaimes2007multimodal, de2021visual}.
With the improvement of ubiquitous perception capabilities brought by various intelligent mobile devices, people expect machines to not only acquire single-model information, but also autonomously complete tasks by combining multimodal context information. 
%With the breakthroughs of these modality-specific technologies in certain fields and the improvement of pervasive perception capabilities brought by various intelligent terminal devices, people expect machines to have not only single-modal perception and computation capabilities, but also autonomously complete tasks of combining multimodal information, especially in the field of combining CV and NLP (i.e., the \textbf{Vision-Language} research area) \cite{aafaq2019video, mogadala2021trends, uppal2022multimodal}. 
Recently, researches of combining language and visual information from their original independent fields (i.e., the \textbf{Vision-Language} research area \cite{ mogadala2021trends, uppal2022multimodal}) have received extensive attention, such as video/image caption generation \cite{ding2019long, ji2021improving}, multimodal machine translation \cite{pham2019found, zhao2022region}, image-text retrieval \cite{chen2020imram, long2022gradual} and visual question answering \cite{norcliffe2018learning, luo2022depth}. These tasks require intelligent systems to unify process, understand the context information in both the visual scenes and textual language, and realize the cross-model reasoning. Specifically, the \textbf{Visual-Context Augmented Dialogue System (VAD)} \cite{nguyen2020efficient, le2020learning} is one of the most fundamental tasks in Vision-Language researches \cite{mogadala2021trends}, which can perceive and understand context information from different modalities in surrounding environments of users (e.g., visual and audio context in the images or videos, textual context in the dialogue history and queries) for realizing harmonious human-machine interaction.

\begin{figure}[t]
\centering
\includegraphics[width=0.9\columnwidth]{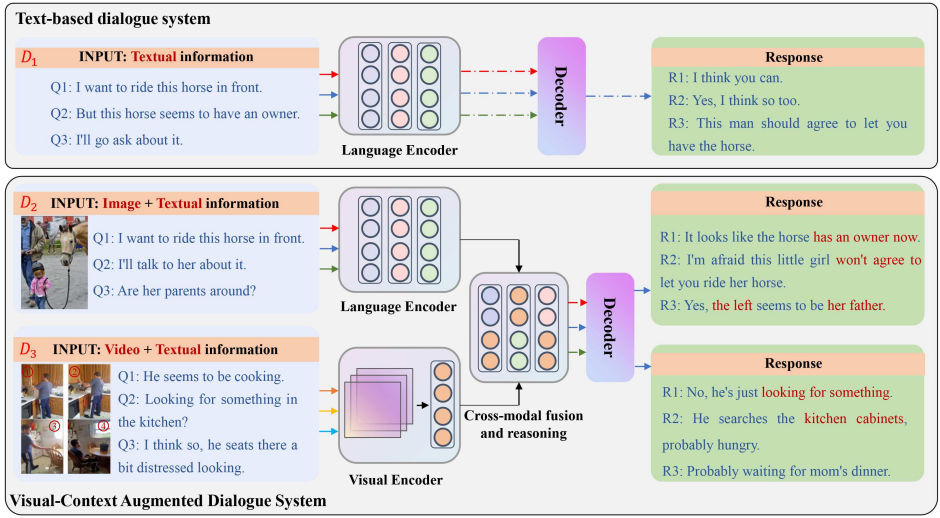}
\caption{Comparison and examples of traditional text-based dialogue system $D_{1}$ and VAD (including image-based dialogue system $D_{2}$ and video-based dialogue system $D_{3}$) from the perspective of input source and feature interaction.}
\label{Dialogue example}
\end{figure}

Compared with traditional text-based dialogue systems, VAD can generate reasonable responses to queries of humans, by not only keeping track of the textual historical dialogue context, but also understanding the visual context information from static images or dynamic videos. By understanding surrounding context through intuitive visual information, VAD is able to be comprehensively aware of interaction demands of users, thus generating more meaningful and engaging responses. Fig.~\ref{Dialogue example} illustrates the comparison between traditional text-based dialogue systems (i.e., $D_{1}$) and VADs (i.e., $D_{2}$ and $D_{3}$). For input sources, the former only can process textual language information (e.g., the pure text input in $D_{1}$), while VAD takes the image or video frames as additional inputs (e.g., the image of \textit{a little girl leading a horse} in $D_{2}$, the video of {a moving man in the kitchen} in $D_{3}$) to perceive visual information as humans. In addition, VAD requires cross-modal fusion and reasoning module to understand the semantic interaction between visual context and textual language information. Without grounding with visual information, text-based dialogue system is more likely to generate fluent but inaccurate response (ignoring some important context information). In dialogue example $D_{1}$, the dialogue agent only produces meaningless affirmative responses to visually relevant queries, which will seriously reduce the users' desire for conversation. 
In contract, VAD has the potential to produce engaging responses that are consistent with visual scenes, thus enhancing the interaction experience. Dynamic motion scenes can also be captured by VAD to reflect real world changes (e.g., the movement flow of the man in $D_{3}$).

%To produce meaningful responses, the dialogue agent must have the ability to perceive and comprehend information from different modalities (visual and audio information in the images or videos, text information in the dialogue history and queries) and integrate the cross-modal information for reasoning as humans \cite{nguyen2020efficient, le2020learning}. Multimodal dialogue system is an essential component of many complex AI systems and applications, and the key technology for realizing natural interaction between humans and machines. For example, in post-disaster rescue scenarios, it is dangerous for rescuers to directly enter the disaster site for investigation. Rescue robots equipped with multimodal dialogue technology can obtain visual information of the disaster site through cameras, while listening to the natural language commands of rescuers, to understand and reason over the visual scene for achieve more efficient on-the-spot investigation.

According to the source of visual information, there are two main research areas in VAD, namely image-based dialogue system \cite{das2017visual, de2017guesswhat, kottur2019clevr} and video-based dialogue system \cite{alamri2019audio, le2021dvd}, as shown in Fig.~\ref{Dialogue example}. In image-based dialogue, visual context information is derived from a single static image, and the relationships and interactions among visual objects are constant. Contrarily, video-based dialogue system needs to reason over visual information from dynamic videos to generate visually reasonable responses. The feature space and semantic structure of videos are rich and more complex than images, because videos involve both spatial (e.g., the appearance and position of objects and their relationships) and temporal dimensions (e.g., the flow of actions or motions across multiple video frames). 
The need for real-time human-machine interaction requires VAD to acquire and understand visual information on mobile devices close to users. However, mobile devices are usually resource-constraint, which are difficult to process complex visual data directly. Consequently, the era of intelligent Internet of Things has put forward new requirements and challenges for VAD.

%Contributed by deep learning models trained on huge amounts of data, such as Convolutional Neural Networks (CNN) \cite{krizhevsky2012imagenet}, Recurrent Neural Networks (RNN) \cite{hochreiter1997long} and Transformer \cite{vaswani2017attention}, researches on image processing tasks and text-based dialogue systems in the single modality have made significant progress. However, multimodal dialogue systems are still in the early stages of research, and some exploratory works have emerged, yet new challenges are raised, summarized as follows.

\begin{itemize}
\item Due to the complexity of spatio-temporal features of visual data, the feature extraction and processing of images and videos require the support of computationally intensive neural models, thus overloading resource-constrained mobile devices. How to efficiently process visual context information to provide timely and efficient interaction services is a challenge.
\item Due to the data heterogeneity between different modalities, there are huge semantic gaps between visual and language context feature spaces. How to effectively perceive unique features under different modalities, further conduct the cross-modal semantic fusion and reasoning for response generation is a big challenge.
\item Due to the phenomenon of visual co-reference, there are plenty of referents or abbreviations in conversations to express linguistic concepts or visual objects that have already been mentioned. How to accurately associate references with visual targets for effective visual reference resolution is another challenge to accomplish complex visual and language reasoning. 
\item Traditional evaluation metrics for text-based dialogue systems are unable to measure whether dialogue agents actually understand the visual information in images or videos. How to reasonably evaluate the quality of VAD is to be solved.
\end{itemize}

%和现有综述的对比
To address the above challenges, this paper aims to comprehensively summarize and outline the development of VAD to inspire research interests. As an emerging research area, to our best knowledge, there is no review available yet that provides an exhaustive summary of this field, while some surveys on related topics are presented. 
%文本对话系统综述
For example, some researches systematically summarize the research on text-based dialogue systems \cite{chen2017survey, huang2020challenges, ma2020survey, ni2021recent}. For harmonious human-machine interaction, Ma \textit{et al.} \cite{ma2020survey} and Ni \textit{et al.} \cite{ni2021recent} focus on the progresses of condition-enhanced dialogue system, such as personalized, context-aware and knowledge-based dialogue system. Huang \textit{et al.} \cite{huang2020challenges} also present the challenges of building open domain dialogue systems and review the approaches to address these challenges. 
%Vision-language综述
Moreover, some surveys focus on research advances in the field of vision-language research \cite{wiriyathammabhum2016computer, baltruvsaitis2018multimodal, uppal2022multimodal, mogadala2021trends}. The early work \cite{wiriyathammabhum2016computer} provides a systematic review of vision-language research from the perspective of applications, such as image captioning and visual retrieval. Baltru{\v{s}}aitis \textit{et al.} \cite{baltruvsaitis2018multimodal} explore core technical challenges in multimodal machine learning (e.g., representation, translation, fusion) and structure works according to the addressed challenges. They usually focus on the alignment and interaction between visual and language modalities, without considering the potential issues of acquiring, perceiving and understanding visual contexts during conversation.  
%Visual-context augmented dialogue system may be considered as a specific subtask in these surveys, without intensive investigated.
In addition, for cross-modal semantic interaction and fusion, some surveys investigate the research progress of cross-media mining \cite{peng2017overview, bhatt2011multimedia, wang2016comprehensive}. Peng \textit{et al.} give an overview of cross-media mining, including the concepts, methodologies, major challenges and open issues, to summarize how cross-modal contexts can be effectively fused for downstream tasks. 
%我们的特点
Compared with these surveys, we provide a systematic and comprehensive review of concepts, challenges and applications in the filed of visual-context augmented dialogue system. We summarize representative researches io the filed of VAD from a systematic perspective, from visual context perception and understanding, to visual and language semantic fusion and reasoning, and reasonable response generation. Furthermore, some thoughts on promising research trends are analyzed to promote the research progress. To sum up, our contributions are summarized as follows.

\begin{itemize}
\item Based on the formalized concept model of Visual-Context Augmented Dialogue System (VAD), we design a generic system architecture, which illustrates the processing and interaction flow of visual and language information in VAD, to provide an overall picture for the summary of research works.
\item We make a profound investigation of the development in VAD by presenting the core contribution of each work, according to the addressed challenges. Besides, we also comprehensively summarize available datasets and evaluation metrics.
\item We further discuss some open issues and promising research trends in VAD, including the cognitive mechanisms of human-machine dialogue under cross-modal dialogue context, knowledge-enhanced cross-modal semantic interaction, etc., to promote the development of the research community. 
\end{itemize}

The rest of this paper is organized as follows. We give the concept model and design a generic system architecture of for VAD in Section \ref{chap:Concept}. Then we detail unresolved challenges and investigate representative works of VAD in Section \ref{chap:works}. In section \ref{chap:dataset}, we summarize available datasets and evaluation metrics followed by open issues and promising research trends in Section \ref{chap:open}. Finally, we conclude this article in Section \ref{chap:conclusion}.

\section{Formalized Concept Model and System Architecture for VAD}
\label{chap:Concept}
By distilling and summarizing the core commonalities of existing VAD works, in this section, we first formalize the concept model for VAD and then present the generic system architecture to visualize the overall processing flow of VAD, including multi-source visual data collection, cloud-edge collaboration-based visual data pre-processing and context extraction, cross-modal information interaction and response generation.

\subsection{Formalized Concept Model for VAD}
Given the query text sequence $Q_{L}=(q_{L}^{1},q_{L}^{2},...,q_{L}^{n})$ with $n$ words, and the historical dialogue set $H$ composed of $(L-1)$ pairs of (question, answer), $H=\{(Q_{1},A_{1}),...,(Q_{L-1},A_{L-1})\}$, where $Q_{i}$ and $A_{i}$ represent the query and response sequence of $i-th$ dialogue turn, the objective of the dialogue system is to generate reasonable responses $A_{L}$. Modern dialogue systems are typically implemented with the end-to-end encoder-decoder structure \cite{sordoni2015neural, serban2016building}. The encoder transforms text sequences into dense vectors to capture the semantic information of the input query and dialogue history, as shown in Eq.\ref{Encoder}, where $Encoder$ may be LSTM \cite{hochreiter1997long}, Transformer \cite{vaswani2017attention} or other neural models.

\begin{equation}
\mathbf{Q_{L}}, \mathbf{H} = \textbf{Encoder}(Q_{L}, H)
\label{Encoder}
\end{equation}

Then the decoder is responsible for producing target responses based on the output of the encoder. There are two kinds of decoders in dialogue system, generative decoder and discriminative decoder.

\textbf{Generative decoder:} Encoded vectors are set as the initial state of the decoder network (e.g., LSTM, Transformer). The output is generated word by word in an auto-regressive manner, and at $i-th$ decoder step, the decoder updates its state vector $s_{i}$ according to previous state vectors and dialogue history, and samples each word $y_{i}$ from a distribution $o_{i}$, as shown in Eq.\ref{Decoder}.

\begin{equation}
y_{i} \sim {o_{i}} = P(y|y_{<i};\mathbf{Q_{L}}, \mathbf{H}) = \textit{softmax}(W_{o}s_{i})
\label{Decoder}
\end{equation}
where $W_{o}$ is the weight matrix. Afterwards, the state vector or decoder is updated as follows.

\begin{equation}
s_{i} = \textbf{Decoder}(s_{i-1}, [\textbf{Att}(\mathbf{Q_{L}};\mathbf{H};s_{i-1});y_{i-1}])
\label{Decoder_state}
\end{equation}
where $\textbf{Att}(\mathbf{Q_{L}};\mathbf{H};s_{i-1})$ represents the attentive information in the input query and dialogue history conditioned on state $s_{i-1}$, and $y_{i-1}$ is the vector of the word generated in $i-1$ decoder step.

\textbf{Discriminative decoder:} Encodings are fed into a $softmax$ decoder to compute dot product similarity with each of the response options in the list of candidate responses $A_{L}=\{A_{L}^{(1)},...,A_{L}^{(n)}\}$, consisting of $n$ possible responses. The response with the highest score will be selected as the final response at current dialogue turn.

Based on the formalization of general text-based dialogue system, we then present the concept model of visual-context augmented dialogue system (VAD).

\begin{myDef}
\textbf{\textit{Visual-Context Augmented Dialogue System (VAD):}} Given the input query $Q_{L}$ and the historical dialogue set $H$, VAD aims to perceive and understand the visual context information from the static image $I$ or dynamic video $V$ consisting of $N_{V}$ frames, $V=(v_{1},v_{2},...,v_{N_{V}})$, and finally generate fluent and reasonable responses to appropriately answer the input query.
\end{myDef}

In addition to the text encoder for learning text information in dialogue context and query, VAD needs the additional visual encoder to transform visual context into feature representations, as shown in Eq.\ref{Visual_Encoder}.

\begin{equation}
\mathbf{I}, \mathbf{V} = \textbf{Vis\_Encoder}(I, V)
\label{Visual_Encoder}
\end{equation}
where $Vis\_Encoder$ may be a pre-trained image classification model (e.g., ResNet \cite{he2016deep} and ResNext-101 \cite{hara2018can}) for extracting frame-level features, a pre-trained object detection model (e.g., Faster R-CNN \cite{ren2015faster}) for extracting object-level features in images, or a pre-trained 3D-CNN model (e.g., I3D \cite{carreira2017quo}, C3D \cite{tran2015learning}) for extracting video features cross multiple frames. 

The \textit{visual context} refers to the information contained in visual scenes (static images or dynamic videos) that people focus on and talk about during conversations. For example, humans in the surroundings, their movements, position relationships between humans and objects are typical visual context that conversations may focus on. After the visual context is transformed into feature representations, the cross-modal fusion and reasoning module captures the semantic interaction between visual and textual language information, to obtain the visual-context enhanced dialogue representation $v^{d}$ for the response decoder. Similarly, the decoder takes $v^{d}$ as additional inputs to produce final response with the generative or discriminative manner. As mentioned above, there are two main research areas in VAD, namely image-based dialogue system \cite{das2017visual, de2017guesswhat, kottur2019clevr} and video-based dialogue system \cite{alamri2019audio, le2021dvd}. Image-based dialogue system aims to enable dialogue agents to communicate with humans grounded in static images utilizing a sequence of dialogue history as context. Furthermore, video-based dialogue system conducts conversations based on the content of videos with tens of seconds, rather than a static image. The video dialog task (i.e., the Audio Visual Scene-Aware Dialog (AVSD) task) is firstly proposed in \cite{alamri2019audio}, which expects agents to generate accurate responses according to the given video, audio, and dialog history. Due to the dynamic nature of the video content, the visual information in the video dynamically changes in both temporal and spatial dimensions, such as the movement and interaction of humans and objects, which brings greater challenges to the understanding of visual scene information.

\subsection{Generic System Architecture for VAD}
\begin{figure}[t]
\centering
\includegraphics[width=1.0\columnwidth]{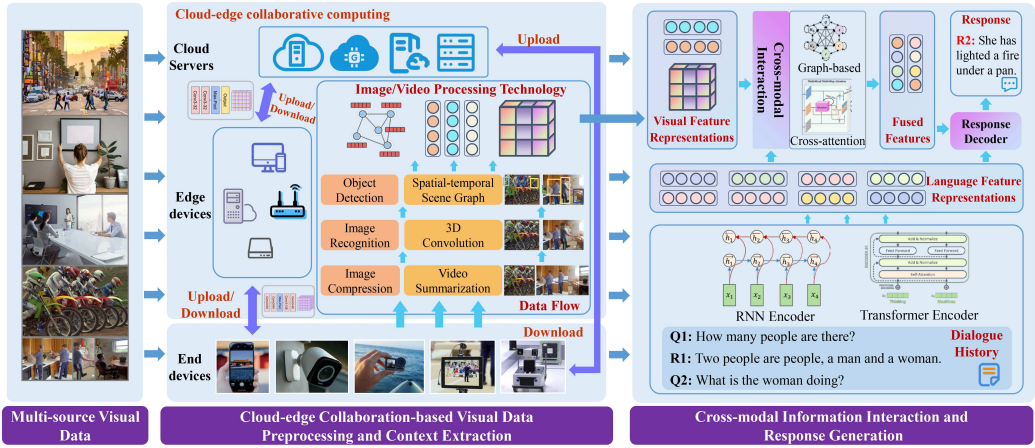}
\caption{The system architecture for VAD, which includes three main modules: multi-source visual data collection, cloud-edge collaboration-based visual data pre-processing and context extraction, and cross-model information interaction and response generation.}
\label{System}
\end{figure}

Based on the definition of VAD, we then present the generic system architecture for VAD to visualize the complete processing flow from visual information collection to reasonable response generation, as illustrated in Fig.~\ref{System}, which consists of the following three submodules:

%in which the key technologies involved are introduced in Table~\ref{tab:key_tech}. The system architecture for VAD is illustrated in Fig.~\ref{System}, which consists of the following three submodules:

\textbf{Multi-source visual data collection.} VAD aims to realize pervasive and visually-grounded human-machine conversational interaction in real surroundings. Equipped with cameras and other sensors, various heterogeneous end devices around users (e.g., smartphones, smart glasses, mobile robots) can perceive and collect visual data in the surrounding environments, including images and videos. As human conversations have frequent topic switching, the visual information that people focus on will also change constantly. The field of view of cameras on different end devices is limited which can only focus on visual scenes within specific spatial scope, making it difficult to completely cover the visual scenes that users are interested in. Meanwhile, there are large gaps in visual data in terms of clarity and image quality collected of different end devices. Consequently, the multi-view fusion-based multi-camera collaboration \cite{ristani2018features, hussain2021comprehensive} is needed to break the field of view limitation of the single device and achieve accurate perception of visual scenes. These visual scenes are then processed and interacted  with dialogue context to respond to the users' conversational needs.

\textbf{Cloud-edge collaboration-based visual data pre-processing and context extraction.}
Due to the spatial-temporal complexity of visual context in images and videos, to provide users with real-time and convenient interaction services, VAD needs to pre-process visual data and extract visual context in advance, supported by advanced image/video processing techniques. Human-human conversations usually focus only on specific regions in images or specific flips in videos. Therefore, removing redundant information in images and videos can effectively improve the processing efficiency of visual information in VAD. Image/video compression \cite{minnen2018joint, ma2019image} and video summarization \cite{zhou2018deep, hussain2021comprehensive} are naturally employed to remove redundant visual information in image and video data. And then, feature extraction technologies, including coarse-grained solutions (e.g., image recognition \cite{dosovitskiy2020image, rawat2017deep}, 3D Convolution \cite{carreira2017quo, yao2019review}) and fine-grained methods (e.g., object detection \cite{sun2021sparse, wang2021salient}, spatial-temporal scene graph \cite{cong2021spatial, ost2021neural}), transform pre-processed image and video data into feature vectors for cross-modal interaction with textual dialogue information.
 
Above visual processing technologies are usually computationally intensive deep neural network models (DNNs), which require sufficient computing and storage resources. However, end devices are usually resource-constraint due to the restricted size and weight, making it challenging to deploy DNNs on them. Cloud-edge collaborative computing \cite{zhou2019edge, ren2019collaborative} emerges to aggregate available resources of cloud, edge, and end devices for collaborative computing, enabling efficient visual data processing. Specifically, model compression technology \cite{he2018amc, deng2020model} reduces the number of parameters of dense layers or reconstructs model structures for lower storage and computation overheads of DNNs, which enables DNNs to be deployed directly on end devices. Moreover, to leverage the powerful computing power of edge devices (e.g., edge servers, converged gateways, smart routers) and cloud servers, model partition technology \cite{kang2017neurosurgeon, wang2021context} splits DNNs into multiple computing modules and distributes different modules to different devices for collaborative computing. The computational and storage burden of visual processing models can be spread across multiple devices to realize the real-time visual data pre-processing and feature extraction. Then visual features will be fed into cross-model information interaction module for response generation.

\textbf{Cross-model information interaction and response generation.}
After extracting visual features in image or video scenes, VAD needs to understand the conversational semantics in dialogue history and the current user query, and reason in visual scenes to generate fluent and reasonable responses. The textual dialogue history and query are firstly encoded into semantic vectors using language encoder (e.g., RNN encoder, Transformer encoder). Afterwards, the cross-model information interaction module (e.g., graph-based reasoning \cite{guo2020iterative}, cross-attention based interaction \cite{le2019multimodal}) is responsible for reasoning on the rich underlying semantic structure between complex visual information and dialogue semantics to capture critical clues for response generation. Finally, the response decoder takes the fused cross-modal features as inputs to produce engaging and informative responses.

\section{Research Challenges and Key Techniques}
\label{chap:works}
Due to the rich and complex visual semantic structures and semantic gaps between the visual and language feature spaces, there are many critical challenges in VAD needed to be addressed. In this section, we first detail these challenges in VAD in section 3.1, and then conduct a comprehensive investigation of representative works to address these challenges in the following subsections (from section 3.2 to section 3.5). Furthermore, some novel training strategies for VAD are also summarized, such as pre-trained vision-language model-based VAD and weighted likelihood estimation based training scheme, as detailed in section 3.6.

\subsection{Research Challenges in VAD}
As an emerging field, researchers have made some explorations in VAD that can achieve a basic understanding of visual scenes and answer the corresponding quires about visual contents. However, there remain massive unresolved challenges in VAD, which will be thoroughly analyzed and summarized in this section.

%三个主要挑战，高效视频处理，跨模态语义推理和融合，视觉指代消解
\textbf{Efficient video processing and understanding.}
VAD needs to extract semantic features in visual scenes from static images or dynamic videos about the visual content of dialogue concerns. Image feature extraction has made exemplary improvements driven by convolutional neural networks (CNNs) \cite{gu2018recent, khan2020survey}. However, as the naturally information-intensive media, video contains hierarchical multi-granularity structure, such as spatial features in specific frame and temporal features in video fragments, which is unable to be processed by 2D CNNs. Consequently, 3D CNNs have been explored to capture fine-grained spatio-temporal features \cite{ji20123d, tran2015learning}. However, the huge computing and storage burden of standard 3D CNNs has become the bottleneck for efficient video processing and understanding, thus reducing the efficiency of time-sensitive human-machine interaction. Meanwhile, human visual attention about visual objects or regions during conversation is usually localized and dynamically changing, and smooth dialogue process can be achieved without perceiving and understanding all possible visual information, indicating that some visual information is redundant for dialogue understanding and response generation. How to efficiently process and understand video information according to the dialogue concerns to enable the semantic interaction with dialogue context and real-time response generation is a critical challenge. The model-based video processing acceleration \cite{liu2020autocompress, feichtenhofer2020x3d} and the content-based video key information selection \cite{li2021exploring, liu2022video} are explored to solve these problems, see details in section 3.2.

\textbf{Cross-modal semantic relation reasoning.}
Due to the data heterogeneity between different modalities, there are huge semantics gaps between visual and language feature spaces. The semantic interaction between dialogue history context and visual information is progressively changing, and relations among various visual objects and entities in historical text are influenced by the current input query, especially in dynamic videos with spatial and temporal variations. Understanding the semantic associations between conversational context and visual information is necessary to reply dialogue queries accurately. Therefore, how to effectively realize the multi-modal representation learning and cross-modal semantic relation reasoning on rich underlying semantic structures of visual information and dialogue context is one of the key challenge. Researches propose to model images or videos and dialogue as the graph structure \cite{zheng2019reasoning, chen2021gog, geng2021dynamic} and conduct cross attention-based reasoning \cite{schwartz2019factor, nguyen2020efficient, chu2021end} to perform fine-grained cross-modal relation reasoning for reasonable responses generation, see details in section 3.3.

\textbf{Visual reference resolution.}
It has been the long-time consensus that humans use coreferences (e.g., short-hands such as pronouns, synonymous phrases) to refer to the same object or referent in language expressions \cite{winograd1972understanding, grice1975logic, yu2016modeling} to improve communication efficiency. As for VAD, there are plenty of referents or abbreviations in conversations to express linguistic concepts and visual objects that have already been mentioned in previous dialogue context. This phenomenon causes dialogue agents to have difficulty in accurately finding corresponding visual targets based on dialogue context, thus preventing the semantic reasoning for generating responses. How to accurately associate references between visual targets and language entities for effective visual reference resolution is another challenge to accomplish complex visual and language reasoning. Associative reference memory \cite{seo2017visual, kottur2018visual} and progressive attention mechanism \cite{niu2019recursive, kang2019dual} are explored for visual reference resolution, see details in section 3.4.

\textbf{Model adaptivity to new scenes}.
Due to the end-to-end supervised learning strategy with the maximum likelihood estimation (MLE) objective function, VAD is difficult to generalize to unseen scenarios, thus generating generic, repetitive and inconsistent responses, specifically in goal-oriented dialogue system which needs to achieve the target dialogue goal. Reinforcement learning (RL) is introduced to explore different dialogue strategies for more effective dialogue modeling \cite{das2017learning, wu2018you, murahari2019improving}. Moreover, the effective training of deep models requires enormous data, which is still relatively scarce in VAD field. Existing datasets for VAD can only cover limited visual scenes, resulting in the failure of trained dialogue agents to adapt to the diverse scenes in physical human-machine interaction scenarios. The idea of pre-train and fine-tune is widely practiced in various NLP tasks and has the potential to solve the issue of data scarcity in VAD \cite{le2020video, li2021bridging}. Researchers have made extensive efforts to address these challenges, as detailed in section 3.5.

%图像和视频的高效处理，仅考虑提升处理效率的工作，通用的特征提取不考虑
%模型层面，降低复杂度
%内容层面，视频摘要

\subsection{Efficient Video Processing and Understanding}
\begin{figure}[t]
\centering
\includegraphics[width=0.9\columnwidth]{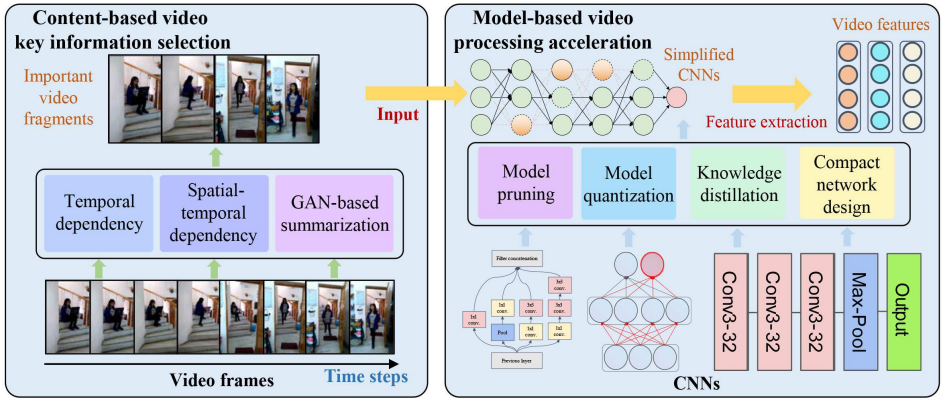}
\caption{Research explorations for efficient video processing and understanding, including the context-based video key information selection and the model-based video processing acceleration.}
\label{Efficient_video}
\end{figure}

For efficiently processing and understanding visual information in videos, researches explore the context-based video key information selection and the model-based video processing acceleration, as shown in Fig.~\ref{Efficient_video}. The former aims to remove redundant information from original videos at the content level, while the latter reduces the complexities of CNNs for extracting video features. Typical works are summarized as follows.

\subsubsection{Content-based video key information selection}
During human-human conversation, we usually focus only on specific objects in specific visual regions, while other visual information is redundant. For improving the video processing efficiency, it is also considerable to remove these redundant information in videos, thus reducing the amount of visual data to be processed and speeding up the video understanding process. Video summarization is the key technology to achieve this target \cite{apostolidis2021video, hussain2021comprehensive}, which selects the most informative and important video frames or fragments to form short synopses that maintain the key information in original videos. Researchers explore to measure the importance of different video frames from different perspectives, which are summarized as follows.

\textbf{Modeling the temporal dependency among frames.} Early video summarization methods formalize summarization as a structured prediction problem which predict frames' importance based on their temporal semantic dependency with other nearby frames. Zhang \textit{et al.} \cite{zhang2016video} firstly investigate to apply LSTM to model variable-range dependencies across video frames based on the determinantal point processes. Further, Ji \textit{et al.} \cite{ji2019video} employ the attentive encoder-decoder framework to assign importance weights to different video frames/shots. The temporal dependencies between frames are captured by the recurrent encoding process to predict importance scores for all frames. To explicitly enhance the diversity of generated summary frames, Li \textit{et al.} \cite{li2021exploring} design the global diverse attention mechanism, which captures temporal relations within paired frames as well as the relations among all pairs, thus considering relations in the whole video. The diverse attention weights are then projected to importance scores for selecting important frames. 

\textbf{Modeling the spatial-temporal dependency among frames/fragments.}
The temporal dependency-based methods can analyze temporal relations between different frames to find relatively redundant frames in videos. However, the spatial characteristics of videos are also critical for determining the importance of frames. To further fuse spatial features in videos, Huang \textit{et al.}  \cite{huang2019novel} propose to extract spatio-temporal information from videos by Capsules Net for generating inter-frames motion curve. The transition effects detection method segments video streams into multiple shots, from which the key frames sequences are selected based on the self attention. Liu \textit{et al.} \cite{liu2022video} introduce the 3D spatio-temporal U-Net to encode spatio-temporal information in videos. Then the RL agent is explored to learn the spatio-temporal latent scores for predicting keeping or rejecting a specific video frame in the video summary. 

\textbf{Enhancing the summarization quality with Generative Adversarial Networks.} To minimize the distance between the generated and the ground-truth summaries, Generative Adversarial Networks (GAN) is employed in video summarization, where the generator is responsible for generating summarizes for given videos and the discriminator quantifies their similarity, thus motivating the generator to improve the summarization quality. Zhang \textit{et al.} \cite{zhang2019dtr} attempt to select key frames with multi-scale temporal context captured by the dilated temporal
relational units and present a discriminator to enhance the information completeness and compactness via the designed loss. Fu \textit{et al.} \cite{fu2019attentive} introduce the attention-based pointer generator to predict the cutting points (the start and end position) for summarization fragment, and the 3D-CNN based discriminator to judge whether a fragment is a generated summarization or from the ground-truth.

\subsubsection{Model-based video processing acceleration}
The most intuitive solution to improve the efficiency of video processing is to elaborate the structure of 3D CNNs to reduce computational requirements and the number of parameters for speeding up the inference process. Various attempts have been made to tune models for efficient video processing \cite{cheng2017survey, deng2020model}, such as model pruning, model quantization, knowledge distillation and compact network design.

\textbf{Model pruning} aims to remove the redundant parameters that do not contribute much to the performance of DNNs \cite{han2015learning, li2016pruning}, thus reducing the storage requirement and the computation of DNNs and improving video processing efficiency. To reduce the redundancy along the spatial/spatial-temporal dimension for efficient 3D CNNs, Guo \textit{et al.} \cite{guo2020multi} propose the Multi-Dimensional Pruning (MDP) framework to simultaneously compress CNNs on both the spatial/spatial-temporal and the channel dimension. An over-parameterized network is constructed by expanding each convolutional layer to multiple branches based on the different spatial/spatial-temporal resolutions. Authors design a gate for each channel to measure its importance, which is used for aggregate different branches. Finally, the branches with lower importance scores are pruned to reduce parameters of models. To accelerate execution of 3D CNNs on edge devices for video processing, Sun \textit{et al.} \cite{sun20203d} introduce a hardware-aware pruning method which leverages the powerful ADMM \cite{liu2020autocompress} and fully adapts to the loop tiling technique of FPGA design. With the support of model pruning technology, the computation and storage of DNNs for video processing can be significantly reduced, thus speeding up the video understanding process.

\textbf{Model quantization} reduces the number of bits of weights and activations in DNNs when stored in physical devices \cite{han2015deep, bulat2021bit} (e.g, reducing from 32-bit floating-point numbers to 16-bit, 4-bit or even with 1-bit), which leads to a significant decrease of the storage size and the number of MAC operations of DNNs. To adapt to the complexity of videos, Sun \textit{et al.} \cite{sun2021dynamic} introduce a input-dependent dynamic quantization framework to assign optimal precision for each frame in videos based on the input complexity for efficient video understanding. A dynamic policy learning network is trained simultaneously with the stand video recognition network to produce the target precision for each frame, with the computational cost loss to both competitive performance and resource efficiency. Furthermore, Lee \textit{et al.} \cite{lee2021qttnet} combine tensor train and model quantization to firstly reduce the number of trainable parameters and then conduct low bit quantization to all parameters to fully lower the memory and time cost produced by 3D CNNs. With lower bit of parameters, the storage requirements, inference time and energy consumption of video processing models can be significantly reduced, making video processing more computational efficient.

\textbf{Knowledge distillation} is a training framework where the bigger cumbersome teacher model is trained on large dataset and then the knowledge embedded in the teacher is transferred to a smaller and lighter student model under the supervision of designed distillation loss \cite{chen2017learning, gou2021knowledge}. Rather than reducing the amount parameters in the training process as above, knowledge distillation directly utilizes lighter and less computational-intensive model to learn teacher model's generalization ability to achieve same/comparable performance. For efficient video inference, Mullapudi \textit{et al.} \cite{mullapudi2019online} propose an online training strategy to specialize low-cost and accurate video semantic segmentation models guided by the learning target provided by the intermittently running teacher model. Differently, Bhardwaj \textit{et al.} \cite{bhardwaj2019efficient} focus on obtaining a compute-efficient student model which process fewer video frames to improve computational efficiency of video classification models. The computationally expensive teacher model processes all frames in video to learn semantic representations of videos and the student model only process a few frames while maintaining the similarity of representations, by minimizing the squared error loss between the representations and the difference between the output distributions between two models. Guided by the teacher model, the light-weight and computational-intensive video processing model can remain relatively high performance for fast and accurate video processing.

\textbf{Compact network design} directly creates computational efficient CNN architectures with computationally lightweight  convolution layers in the model design stage, such as SqueezeNet \cite{iandola2016squeezenet}, MobileNet \cite{howard2017mobilenets} and ShuffleNet \cite{zhang2018shufflenet}. Specifically, for computation and memory efficient video recognition, Kondratyuk \textit{et al.} \cite{kondratyuk2021movinets} design Mobile Video Networks (MoViNets) with three steps to improve improve computational efficiency while reducing the memory usage of 3D CNNs. The  neural architecture search (NAS) is firstly employed to discover efficient and diverse CNN architectures, and then the stream buffer processes videos in small consecutive subclips to save memory consumption. Finally, a simple ensembling technique is proposed to improve the slightly lost accuracy in the second stage, thus achieving accurate and efficient video processing. Feichtenhofer \cite{feichtenhofer2020x3d} choose to progressively expand a tiny based 2D network into a spatiotemporal one for video processing by expanding possible axes, such as the temporal duration and the frame rate. By expanding only one axis at a time, the resultant architecture is trained and validated for selecting the desired architecture that achieves the best computation/accuracy trade-off.

%分类：
%1. Relations reasoning cross modalities->Graph based reasoning methods, Cross attention-based methods
%2. Multimodal co-reference resolution
%3. RL
%4. Pretrainig
%5. Novel training

%Furthermore, Massiceti \textit{et al.} \cite{massiceti2018visual} argues that Visual dialogue can be modeled without considering history or Visual information.A  dedicated response \cite{das2019response}rebutted that there is still need to obtain advanced feature representations of both modalities, complemented by appropriate evaluation metrics.Some research also proved that, for example,Agarwal \textit{et al.}  \cite{agarwal2020history}similarly demonstrated that fine-grained coding of dialog histories can further improve model performance.However, the challenges and controversies of visual dialogue are not just about data processing. According to different research ways, we categorize the research on visual dialog into four main points: multimodal input and processing, reinforcement learning-based approaches, based on the pre-training-fine-tuning paradigm and other innovative strategies, which will be presented in the next subsections.

%合并image和video的工作，
\subsection{Cross-modal Semantic Relation Reasoning}
VAD is an ongoing conversation between interlocutors about the visual information in images or videos. As the conversation progresses, the relations between various visual objects and conversation goals or concerns are dynamically shifted. Therefore, relations between textual and visual modalities are critical for reasoning and understanding rich semantic information among cross-modal entities. In this section, we make an investigation of cross-modal semantic relation reasoning for VAD, as summarized in Table~\ref{tab:Cross_modal_reasoning}.

% \usepackage{multirow}
\begin{table}[]
\caption{A summary of representative works of cross-modal semantic relation reasoning in VAD}
\label{tab:Cross_modal_reasoning}
\centering
\begin{tabular}{|c|c|m{1.5cm}<{\centering}|m{7.8cm}<{\centering}|}
\hline
\textbf{Solution} &
  \textbf{Type} &
  \textbf{Work} &
  \textbf{Description} \\ \hline
\multirow{12}{1.6cm}{Graph-based Reasoning} &
  \multirow{7}{1.3cm}{Image-based dialogue} &
  Zheng \textit{et al.} \cite{zheng2019reasoning} &
  Formalizing image-based dialogue as inference in a graphical model to infer unobserved answer with GNN. \\ \cline{3-4} 
 &
   &
  Guo \textit{et al.} \cite{guo2021context} &
  Encoding joint visual and textual semantic features in one graph and iteratively performing graph attention. \\ \cline{3-4} 
 &
   &
  Chen \textit{et al.} \cite{chen2021gog} &
  Relation-aware graph-over-graph network to reason relations within and among different modality graphs. \\ \cline{3-4} 
 &
   &
  Zhao \textit{et al.} \cite{zhao2021skanet} &
  Incorporating commonsense knowledge to construct the image-dialogue graph. \\ \cline{2-4} 
 &
  \multirow{2}{1.3cm}{Video-based dialogue} &
  Geng \textit{et al.} \cite{geng2021dynamic} &
  Dynamic scene graph representation to encode spatial and temporal semantic information in videos. \\ \cline{3-4} 
 &
   &
  Kim \textit{et al.} \cite{kim2021structured} &
  Gradually neighboring graph attention to capture local-to-global spatio-temporal dynamics. \\ \hline
\multirow{10}{1.6cm}{Cross Attention-based Reasoning} &
  \multirow{6}{1.3cm}{Image-based dialogue} &
  Gan \textit{et al.} \cite{gan2019multi} &
  Inferring the answer progressively through multiple dual attention-based reasoning steps to find and update visual and textual clues. \\ \cline{3-4} 
 &
   &
  Guo \textit{et al.} \cite{guo2019dual, guo2020textual} &
  Imposing textual features on global and local visual features respectively and then exploring mutual correlation through a dual crossing attention. \\ \cline{3-4} 
 &
   &
  Schwartz1 \textit{et al.} \cite{schwartz2019factor} &
  Aggregating information from modalities in the graph based attention framework using message passing. \\ \cline{2-4} 
 &
  \multirow{2}{1.3cm}{Video-based dialogue} &
  Hori \textit{et al.} \cite{hori2019end} &
  Feeding visual and textual features into the multimodal attention to achieve cross-modal reasoning. \\ \cline{3-4} 
 &
   &
  Le \textit{et al.} \cite{le2020bist} &
  Performing bi-directional spatial and temporal reasoning to capture higher-resolution features for input queries. \\ \hline
\end{tabular}
\end{table}

\subsubsection{Graph-based Semantic Relation Reasoning}
Graph neural networks (GNN) \cite{kipf2016semi, velivckovic2018graph, zhang2019heterogeneous} have sparked a tremendous interest at various deep learning tasks \cite{wang2019neighbourhood, gu2019scene}. The core idea of these works is to model the underlying structures of input data as the graph structure and combine the graphical structural representations with neural networks, which is quite powerful for reasoning-style tasks. Due to the brain-like reasoning process and a more concise way to present dependence information, GNN has achieved outstanding performance in various Vision-Language tasks \cite{teney2017graph, li2019relation, wang2021structured}. Consequently, researchers have embarked on exploring the potential of GNNs in VAD. 

For example, Zheng \textit{et al.} \cite{zheng2019reasoning} explicitly formalize image-based dialogue system as inference in a graphical model to reason over underlying semantic dependencies of dialogue history and image entities. The Expectation Maximization algorithm is used to infer both the observed nodes (entities in dialogue history) and the missing node value (desired answers) and a differentiable GNN solution approximates this process. To model the progressively changing relationships among the objects in the image and the dialogue history, Guo \textit{et al.} \cite{guo2020iterative} introduce the Context-Aware Graph (CAG) neural network, in which each node corresponds to a joint semantic feature, including both visual object feature and textual history feature. The graph structure is iteratively updated using an adaptive top-K message passing mechanism conditioned on input query, and the graph attention is performed on all nodes to infer the response. Furthermore, the visual-aware knowledge distillation mechanism \cite{guo2021context} is proposed to remove the noisy historical context feature on the basis of CAG. Jiang \textit{et al.} \cite{jiang2020kbgn} present to build cross-modal information bridges for capturing the underlying text and the vision relation in fine granularity. The Knowledge Retrieval module adaptively selects relevant information in different modalities for answer prediction.

Although above works have employed graph-based structure, their models still lack explicitly capturing complex relations within visual information or textual contexts. Chen \textit{et al.} \cite{chen2021gog} produce the graph-over-graph network (GoG), which consists of three cross modalities graph to capture relations and dependencies between query words, dialogue history and visual objects in image-based dialogue. Then the high-level representation of cross-modal information is used to generate visually and contextually coherent responses. By leveraging a new structural loss function, Sparse Graph Learning (SGL) \cite{kang2021reasoning} learns the semantic relationships among dialog rounds and predicts sparse structures of the visually-grounded dialog. To identify multiple possible answers, the Knowledge Transfer (KT) method is also proposed to extract the soft scores of each candidate answer. Only the internal knowledge implicit in the image and dialogue history graph is insufficient to reason complex relations of visual conversations. To solve this, Zhao \textit{et al.} \cite{zhao2021skanet} introduce the commonsense knowledge to construct semantic relevance among objects and dialogue history for more logical reasoning. After the attention-based fusion of cross-modal features, the structured knowledge graph is constructed by matching objects and textual entities from ConceptNet \cite{speer2017conceptnet}. Then the GCN is employed to extract the relations context for generating responses.

%video
The objects and their spatial relationships in static images are fixed, so the modeling of the semantic information of images and the dialogue context can be achieved using a static graph structure. However, the temporal and spatial variations in videos (e.g., the movement of humans, the interaction between objects) require the dynamic graph structure for fine-grained semantic relation modeling. Geng \textit{et al.} \cite{geng2021dynamic} introduce the spatio-temporal scene graph to encode the video sequences, which extracts objects and their relation using pre-trained Faster R-CNN to build a scene graph \cite{johnson2015image} for every (temporally-sampled) frame. Then the intra-frame reasoning module and inter-frame aggregation module are applied on the scene graph to encode semantics of the video from spatial and temporal dimensions with graph attention. Finally, the multi-head Transformer fuses cross-modal semantic features to generate responses. Instead of only representing videos as graph structures, Kim \textit{et al.} \cite{kim2021structured} construct the visual graph and also the textual graph with words as nodes. The textual and visual co-reference resolution are performed on two graphs with graph attention, and local-to-global dynamics are captured via gradually neighboring graph attention. Similarly, the framework of Reasoning Paths in Dialogue Context (PDC) learns a reasoning path that traverses within video semantics and dialogue history in the graph structure, to propagate visual information and contextual cues related to current questions. 

\subsubsection{Cross Attention-based Semantic Relation Reasoning}
Attention mechanisms are widely explored in various deep learning tasks, especially in Vision-Language tasks, such as video question answering \cite{zhang2021multimodal, rahman2021improved} and image caption generation \cite{yan2021task, lu2021chinese}, to identify critical information in different input modalities for information interaction and semantic fusion. For better acquiring and reasoning over cross-modal semantic correlations to generate reasonable responses, the cross attention mechanism has been adapted to VAD, and demonstrated compelling results.

Gan \textit{et al.} \cite{gan2019multi} propose the Recurrent Dual Attention Network (ReDAN) which infers the answer progressively through multiple reasoning steps. In each reasoning step, the cross attention between modalities is adopted to find and update visual and textual clues that are most relevant to the question, to obtain the multimodal context vector for answer prediction. For different questions, local and global visual attention maps may result in different responses related to textual semantics. To utilize both global image-based and local object-based visual features, Guo \textit{et al.} \cite{guo2019dual, guo2020textual} introduce two visual reasoning steps. The first step imposes textual semantic features on global and local visual features respectively and the second step explores mutual correlation through a dual crossing attention between global and local visions. The answer is inferred by a multimodal semantic fusion module to fuse salient semantics in rich context acquisition. Previous methods typically leverage single-hop reasoning or single-channel reasoning, which is intuitively insufficient for handling complex cross-modal reasoning.. Facing this problem, Chen \textit{et al.} \cite{chen2020dmrm} present the Dual-channel Multi-hop Reasoning Model which maintains a dual channel to obtain attended image features relying on question and history features, and attended dialog history features relying on question and image features, by a multi-hop reasoning process. The attended features are fed into multimodal attention enhanced decoder to generate more accurate responses.

Go further, for video-based dialogue system, the temporal and spatial features are needed to be considered to comprehensively capture the dynamic video information. 
Early practices obtains temporal and spatial features by sampling some frames from the video and extracting visual features for each frame using pre-trained image recognition models (e.g., VGG19) \cite{schwartz2019simple, hori2019end}. Then visual and dialogue context features are fed into multimodal attention module to achieve cross-modal semantic reasoning for answering the input query. To learn a more strong joint-aligned semantic space between the video and textual dialogue, Sanabria \textit{et al.} \cite{sanabria2019cmu} and Pasunuru \textit{et al.} \cite{pasunuru2019dstc7} propose the hierarchical and bidirectional attention for deeper semantic interaction. For simulating the process of human perception of visual context, BiST \cite{le2020bist} performs bi-directional reasoning, including spatial $\to$ temporal reasoning to infer spatial information on target objects in the video and temporal $\to$ spatial reasoning to focus temporal information on target video segments, to capture higher-resolution video features for input queries.

The multi-head self-attention mechanism in Transformer is also explored to capture the complex dependencies of long-term sequence information (e.g., the time dependencies in long videos and multi-turn dialogue history). Multimodal Transformer Network (MTN) \cite{le2019multimodal} applied multi-head attention across multiple modalities (visual, audio and textual dialogue history) repeatedly to capture complex sequential information over cross-modal information. Above models utilize the attention mechanism to attend on all utilities (i.e., the input query, the visual information and the history), making it computationally and conceptually challenging. To address this issue, Schwartz1 \textit{et al.} \cite{schwartz2019factor} develop a general factor graph based attention mechanism, which uses a graph based formulation to represent the attention framework. The nodes in the graph represent different utilities and factors model their interactions, and a message passing like procedure aggregates information from modalities to infer the final answer. When the dialogue grows long, building the factor graph is also computationally inefficient. Light-weight Transformer for Many Inputs (LTMI) \cite{nguyen2020efficient} is produced to deal with all interactions between multiple utilities. The input feature space of multi-head attention is divided to subspaces mechanically and interactions of multiple utilities are computed to one utility to retain sufficient representational power with a much fewer number of parameters.

\begin{figure}[t]
\centering
\includegraphics[width=0.9\columnwidth]{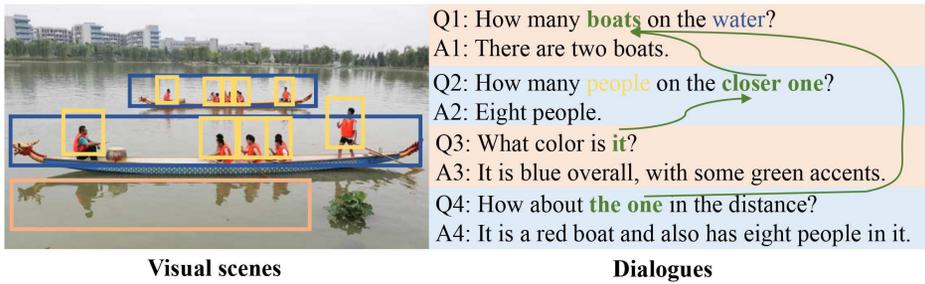}
\caption{An dialogue example with visual reference resolutions, where the concerned entity \textit{boat} in the conversation is referred to by three phrases, \textit{the closer one}, \textit{it} and \textit{the one in the distance}, respectively.}
\label{visualcore_exam}
\end{figure}

\subsection{Visual Reference Resolution}
The phenomenon of co-reference in human language makes coreference resolution, which aims at recognizing noun phrases or pronouns representing of the same entity, be a fundamental research field in NLP communities \cite{soon2001machine, yu2016modeling, stylianou2021neural}. As for VAD, the problem evolves into visual reference resolution, which is the key component for visual dialogue agents to explicitly and accurately locate related objects in images and videos guided by textual entities in the input query and dialogue history, for intensive understanding of visual information. Fig.~\ref{visualcore_exam} shows an dialogue example with visual reference resolutions. The concerned entity that the entire dialogue focused on is \textit{boat}, which is referred to by various pronouns (\textit{the closer one}, \textit{it}, \textit{the one in the distance}) in the three subsequent rounds of the conversation. To response these questions accurately, VAD needs to conduct visual reference resolution to completely understand the dialogue context and reason over visual information. In this section, we summarize researches contributed to visual reference resolution for VAD and representative researches to address challenges in the subsequent subsections are summarized in Table~\ref{tab:Other_work}.

Seo \textit{et al.} \cite{seo2017visual} employs an associative memory into the attention mechanism to obtain a visual reference for an ambiguous expression. Through an associative attention memory, two types of intermediate attentions, tentative and retrieved ones, are designed for capturing related visual region based on the current question and previous attention information, to modeling the sequential dependency of visual reference. For more fine-grained word-level visual coreference resolution, Kottur \textit{et al.} \cite{kottur2018visual} propose to store track, and locate entities explicitly, which accomplishes interpretable visual co-reference inference at the word level rather than sentence level. Considering that humans usually only review the topic-related dialogue history to achieve accurate visual co-reference, Niu \textit{et al.} \cite{niu2019recursive} expect the dialog agent to selectively refer the dialogue history like humans during the conversation. Recursive Visual Attention (RvA) is proposed to ground the visual context based on the input query firstly. If failed, the visual attention will be refined by recursively reviewing the topic-related dialog history until the answer from the visual information can be confidently referenced. Similarly, Kang \textit{et al.} \cite{kang2019dual} address the visual reference resolution as humans two-stage process. The REFER module learns to retrieve relevant dialogue history for linguistical coreference resolve to questions and then the FIND module performs visual grounding via bottom-up attention mechanism to improve the quality of generated responses.

The above works all implicitly attend to spatial or object-level image features, which will be inevitably distracted by unnecessary visual content. To address this, Chen \textit{et al.} \cite{chen2021multimodal} establish specific mapping of objects in the image and textual entities in the input query and dialogue history, to exclude undesired visual content and reduce attention noise. Additionally, the multimodal incremental transformer integrates visual information and dialogue context to generate visually and contextually coherent responses.

% \usepackage{multirow}
\begin{table}[]
\caption{A summary of representative works according to resolved problems.}
\label{tab:Other_work}
\centering
\begin{tabular}{|c|m{1.6cm}<{\centering}|m{9cm}<{\centering}|}
\hline
\textbf{Problem} &
  \textbf{Work} &
  \textbf{Description} \\ \hline
\multicolumn{1}{|c|}{\multirow{5}{2cm}{Visual Reference Resolution}} &
  Seo \textit{et al.} \cite{seo2017visual} &
  Capturing related visual region through an associative attention memory. \\ \cline{2-3} 
 &
  Niu \textit{et al.} \cite{niu2019recursive} &
  Selectively referring dialogue history to refine the visual attention until referencing the answer. \\ \cline{2-3} 
 &
  Chen \textit{et al.} \cite{chen2021multimodal} &
  Establishing mapping of visual object and textual entities to exclude undesired visual content. \\ \hline
\multicolumn{1}{|c|}{\multirow{8}{2cm}{Visual-based Dialogue Strategies Optimization}} &
  Das \textit{et al.} \cite{das2017learning} &
  Generalizing image dialogue as a RL-guided cooperative game between two dialog agents. \\ \cline{2-3} 
 &
  Wu \textit{et al.} \cite{wu2018you} &
  Enhancing response generator with discriminator by RL reward. \\ \cline{2-3} 
 &
  Shukla \textit{et al.} \cite{shukla2019should}&
  Maximizing the information gain while asking questions with a RL paradigm for explicit dialogue goals. \\ \cline{2-3} 
 &
  Zhou \textit{et al.} \cite{zhou2019building} &
  Optimizing dialogue policies by RL and ensuring natural responses with supervised learning. \\ \hline
\multicolumn{1}{|c|}{\multirow{3}{2cm}{Pre-trained Vision Language Model-based VAD}} &
  Murahari \textit{et al.} \cite{murahari2020large} &
  Pre-traning on vision-language datasets to learn visually-grounded language representations. \\ \cline{2-3} 
 &
  Wang \textit{et al.} \cite{wang2020vd} &
  Training unified Transformer encoder initialized by BERT with two visual training objectives. \\ \cline{2-3} 
 &
  Le \textit{et al.} \cite{le2020video} &
  Utilizing GPT-2 to capture cross-modal semantic dependencies. \\ \hline
\multicolumn{1}{|c|}{\multirow{5}{2cm}{Unique Training Schemes-based VAD}} &
  Zhang \textit{et al.} \cite{zhang2019generative}&
  Assigning different weights to training samples to enhance the diversity of responses. \\ \cline{2-3} 
 &
  Jiang \textit{et al.} \cite{jiang2020dualvd} &
  Simulating Dual-coding theory of human cognition to adaptively find query-related information from the image. \\ \cline{2-3} 
 &
  Testoni \textit{et al.} \cite{testoni2021looking} &
  Asking questions to confirm the conjecture of models about the referent guided by human cognitive literature. \\ \hline
\end{tabular}
\end{table}

\subsection{Visual-based Dialogue Strategies Optimization}
The aforementioned works cast VAD into to one of supervised learning problem based on the sequence to sequence structure, whose objective is to maximize the distribution of the response given dialogue history and visual information. Due to the vast dialogue action space, existing dialogue datasets can cover only a small subset of conversation scenarios, making dialogue agents are difficult to generalize to unseen scenarios. Naturally, the dialogue generation should be an interactive agent learning problem, which needs to learn an optimal strategy to efficiently accomplish the intrinsic goal of conversations to steer the conversation in unseen conversation scenarios. To deal with the above problem, RL is introduced to exploring different dialog strategies \cite{das2017learning, strub2017end} for more effective, consistent responses and a higher level of engagement of dialogue agents in the conversation.

For the first time, Das \textit{et al.} \cite{das2017learning} generalize the image-based dialogue beyond the supervised learning by posing it as a cooperative ``image guessing'' game between two dialog agents, with RL to learn dialogue polices of dialogue agents. Through designing a questioner bot (Q-BOT) and an answerer bot (A-BOT), the RL mechanism makes the automatic emergence of grounded language and communication among dialogue agents without human supervision. Meanwhile, Strub \cite{strub2017end} present a global architecture with RL to optimize visually task-oriented dialogues, based on the policy gradient algorithm. Murahari \textit{et al.} \cite{murahari2019improving} improve the problem of repetitive interaction between Q-BOT and A-BOT on image-independent information. The smooth-L1 penalty is introduced to penalize similarity in successive state vectors, thus motivating Q-BOT to ask a wide variety of questions and A-BOT to focus on more visual information. To improve the problem that commonly used policy-based RL agents often end up focusing on simple utterances and suboptimal policies, Zhao\cite{zhao2018improving} propose temperature-based extensions for policy gradient methods which encourages exploration with different temperature control strategies to generate more convincing responses. Inspired by GAN, Wu \textit{et al.} \cite{wu2018you} jointly train two submodules with RL: a sequence generative model to generate responses based on the dialogue history and image, and a discriminator to encourage generator to generate more human-like dialog by distinguishing generated responses and human real ones. The output of the discriminator is seen as the reward to optimize the training process.

Motivated by linguistics and psychology for modelling human conversation, Shukla \textit{et al.} \cite{shukla2019should} maximize the information gain of dialogue agent while asking questions with a RL paradigm for explicit dialogue goals. Previous RL-based approaches specialize in achieving an optimal dialog policy to complete the external dialogue goal, but inevitably compromise the quality of generated responses. Facing this problem, Zhou \textit{et al.} \cite{zhou2019building} separate dialogue strategy learning from language generation by means of alternate training. The dialogue policies are optimized by RL and additional supervised learning mechanism is supplemented to ensure naturalness of responses with synchronous optimization. To explore the impact of imperfect dialogue history on dialogue strategies, Yang \textit{et al.} \cite{yang2019making} intentionally impose wrong answers in the history and see the model future behavior. The History-Advantage Sequence Training (HAST) comprehensively integrates the dialogue history with co-attention modules for visual context in history encoding and one history-aware gate. For more efficient training dialogue models with RL, the probabilistic framework is introduced \cite{chang2019learning} with an Information Gain Expert and a Target Posterior Expert, which provide virtually unlimited expert demonstrations for pre-training the questioner and will be refined for a even better policy with RL.

\subsection{Novel Training Strategies for VAD}
In addition to above studies to address these typical challenges, researchers propose some novel training strategies to improve the performance of VAD, which will be introduced in this section.

\subsubsection{Pre-trained Vision-Language Model-based VAD}
With the emergence of Transformer, the pre-training and fine-tuning paradigm has been extensively explored in both CV and NLP areas. By pre-training neural network models on large-scale unstructured text or visual data to learn universal implicit knowledge for natural language or visual understanding, and fine-tuning models with task-specific data, excellent performance can be achieved. For natural language understanding and generation, numerous variants of BERT \cite{devlin2019bert} and GPT \cite{radford2018improving} accomplish impressive breakthroughs in various tasks \cite{liu2019text, sun2019utilizing, wu2020tod}. Furthermore, ViLBERT \cite{lu2019vilbert} extends the idea of pre-training to the Vision-Language areas for unified modeling and understanding of natural language and visual information. Through the unified pre-training process, visual language pre-training specializes in modeling cross-modal semantic spaces and aligning semantics of vision and language to provide sufficient prior knowledge for subsequent tasks \cite{su2019vl, yu2021ernie, radford2021learning}. Researchers have adopted the pre-training paradigm in VAD, and achieved outstanding performance.

ViLBERT \cite{lu2019vilbert} firstly introduce two separate streams for visual and language processing to accommodate the different needs of each modality and the interaction between modalities through the co-attentional transformer layers. After pre-training, ViLBERT can meet downstream tasks easily by adding a simple classifier. For specialized in VAD, Murahari \textit{et al.} \cite{murahari2020large} adapt ViLBERT for multi-turn visually-grounded conversations. The the large-scale Conceptual Captions dataset \cite{sharma2018conceptual} and Visual Question Answering (VQA) \cite{antol2015vqa} dataset are chosen for pre-traning to learn comprehensive visually-grounded language representations. Considering the multi-turn dialogue history input, authors also present additional segment embeddings to represent questions-answer pairs. Without pre-training on external vision-language data, VD-BERT \cite{wang2020vd} fed the image and dialogue history into unified Transformer encoder initialized with BERT. The two training objectives, Masked Language Modeling (MLM) and Next Sentence Prediction (NSP), are modified to consider visual information when predicting the masked tokens or the next answer. Tu \textit{et al.} \cite{tu2021learning} produce novel Oracle, Guesser and Questioner models built on pre-trained VilBERT. The two-way background/target fusion mechanism is introduced to understand both intra and inter-object question in the Oracle model. The Guesser model uses a state-estimator to realize the potential of VilBERT in single-turn referring expression comprehension. For the Questioner, the shared state estimator from pre-trained Guesser guides the question generator.
%The multimodal pre-training model on visual and language still needs to be explored, such as further considering the inclusion of dynamic information such as video modality to simulate a more realistic and natural intelligent interaction world.

The pre-trained model-enhanced video-based dialogue systems have also attracted the attention of researchers. VideoBERT \cite{sun2019videobert} is the first attempt to pre-train video language tasks by performing self-supervised multi-task learning on a large amount of video, audio and text training data, providing a heuristic reference for extending pre-trained models to video-based dialogue systems. Le \textit{et al.} \cite{le2020video} formulate video-based dialogue tasks as a sequence-to-sequence task, and fine-tune large pre-trained GPT-2 model to capture dependencies across multiple modalities. The video features of multiple frames are placed as temporal sequence and combined with input text sequence (i.e., the dialogue history and response) to train the model with masked multi-modal modeling objectives.

\subsubsection{Unique Training Schemes-based VAD}
The traditional training paradigm of neural dialog models is maximum likelihood estimation (MLE), resulting in a tendency to generate frequent and generic responses with high frequency of occurrence in the dataset, such as ``I don't know'' and ``I can't tell''. The tendency to produce repetitive or overly simple and intuitive responses are abottleneck problems facing by anthropomorphic dialogue systems. To obtain more informative responses, Lu \textit{et al.} \cite{lu2017best} introduce the adversarial training to train a generative dialogue model $G$ with self attention, which receives gradients from discriminative dialogue model $D$ to encourage more diverse and informative responses. Similarly, Zhang \textit{et al.} \cite{zhang2019generative} propose the weighted likelihood estimation (WLE) based training scheme. Training samples in the dataset are assigned different weights in the training process according to positive response as well as the negative ones of them, to enhance the diversity of responses.

As a typical simulation of human interaction, the human cognitive processes also inspire the training of VAD. Motivated by Dual-coding theory \cite{paivio2013imagery} of human cognition, Dual Encoding Visual Dialogue (DualVD) model \cite{jiang2020dualvd} adaptively finds query-related information from the image through intra-modal visual features and inter-modal visual-semantic knowledge semantics. Based on a beam search re-ranking algorithm, Testoni \textit{et al.} propose Confirm-it \cite{testoni2021looking}, which asks questions to confirm the conjecture of models about the referent with human cognitive literature on information search and cross-situational word learning. To explore the ability of AI dialogue agents to both ask questions and answer them as humans, researchers have made preliminary explorations. Jain \textit{et al.} \cite{jain2018two} develop a unified deep net architecture for both visual question answering and question generation. Further, FLIPDIAL model \cite{massiceti2018flipdial} adopts CNN to encode dialogue history for capturing dialogue context and conditional VAE to learn the generative model, to generate both responses and entire sequences of dialogue (query-answer pairs) which are diverse and relevant to the image.

\section{Datasets and evaluation metrics for VAD}
\label{chap:dataset}
\subsection{Visual-context Augmented Dialogue Datasets}
Massive data is undoubtedly an essential factor for training deep neural networks-based VAD. To promote researches in this field, many large-scale and high-quality datasets have been released, summarized in Table~\ref{tab:datasets}.

\begin{table}[htbp]
  \centering
  \caption{A review of available datasets.}
  \label{tab:datasets}
 
    \begin{tabular}{m{1.9cm}<{\centering}m{2.2cm}<{\centering}m{8.4cm}<{\centering}}
    \toprule
    \textbf{Datasets} & \multicolumn{1}{c}{\textbf{Type}} & \multicolumn{1}{c}{\textbf{Description}} \\
    \midrule
    VisDial \cite{das2017visual} & Image-based Dataset & The first large-scale image-based dialogue dataset, including question-answer pairs about images. \\
    \midrule
    GuessWhat?! \cite{de2017guesswhat} & Image-based Dataset & A dataset based on a two-player guessing game, where agents understand image scenes by asking questions. \\
    \midrule
    CLEVR-Dialog \cite{kottur2019clevr} & Image-based Dataset & A diagnostic dataset for studying multi-round reasoning and visual coreference resolution in image-based dialogue. \\
    \midrule
    Image-Chat \cite{shuster2020image} & Image-based Dataset & A dialogue dataset in which speakers are required to play roles with specific emotional mood or conversation style. \\
    \midrule
    OpenViDial \cite{meng2020openvidial} & Image-based Dataset & An open-domain dialogue dataset with image contexts extracted from movies and TV series. \\
    \midrule
    Dodeca \cite{shuster2020dialogue} & Image-based Dataset & A large dialogue dataset collection containing 12 subtasks, promoting the conversation ability of dialogue agents. \\
    \midrule
    AVSD \cite{alamri2019audio}  & Video-based Dateset & The first video-based dialogue dataset, with conversations discussing the videos about indoor human activities. \\
    \midrule
    DVD \cite{le2021dvd}   & Video-based Dateset & A diagnostic video-grounded dialogue dataset containing annotations for different types of reasoning over videos. \\
    \bottomrule
    \end{tabular}%
  \label{tab:addlabel}%
\end{table}%

%However, in open domain dialogue systems, visual grounded dialogue datasets are mostly collected from major web platforms, which inevitably have a large amount of errors and noise. In 2020, Qi \textit{et al.} \cite{qi2020two} clearly pointed out the hidden causality of visual dialogue models and data and proposed two improvement principles to enhance the generalization ability of the models. Nevertheless, high-quality dialogue data is still relatively scarce. How to construct high-quality and large-scale benchmark dialogue data set and make effective use of other advanced methods such as data enhancement technology in open domain dialogue system is also the bottleneck challenge of the current dialogue system and the direction of future efforts.

\subsubsection{Image-based Dialogue Datasets.}
For training image-based dialogue systems, researchers collect image-paired multi-turn dialogue data from social media platforms or crowdsourcing Platforms.

\textbf{VisDial.} VisDial\footnote{\url{https://visualdialog.org/}} \cite{das2017visual} is the first large-scale image-based dialogue dataset, which contains a total of 1.4 million conversational question-answer pairs and 140,000 images. Each conversation in it includes 10 question-answer pairs discussing the corresponding image from the COCO dataset \cite{lin2014microsoft}. The image-based dialogue task is formally presented after VisDial, which requires dialogue agents to communicate naturally with humans about visual content.

\textbf{GuessWhat?!.}  De \textit{et al.} \cite{de2017guesswhat} collect a large-scale dataset, named GuessWhat?!\footnote{\url{https://guesswhat.ai/download}}, based on a two-player guessing game that included 150,000 games played by humans, with a total of 800,000 visual question-answer pairs on 66K images. This dataset requires dialogue agents to deeply under the image scene (e.g., spatial reasoning and language grounding) by asking a sequence questions, to promote the immediate interaction capability of dialogue systems.

\textbf{CLEVR-Dialog.} CLEVR-Dialog\footnote{\url{https://github.com/satwikkottur/clevr-dialog}} \cite{kottur2019clevr} is a large-scale diagnostic dataset for studying multi-round reasoning and visual coreference resolution in image-based dialogue. Specifically, authors construct a dialog grammar based on the image scene graphs from the CLEVR image dataset \cite{johnson2017clevr}. Overall, CLEVR-Dialog contains five instances of 10-round dialogues involving about 8,500 CLEVR images with a total of 4.25 million question-answer pairs.

\textbf{Image-Chat.} Shuster \textit{et al.} \cite{shuster2020image} present a human-human conversations dataset, Image-Chat\footnote{\url{http://parl.ai/projects/image_chat}}, in which speakers are required to play roles with specific emotional mood or conversation style, making conversations more natural and emotionally rich. The overall dataset consists of 202,000 conversations on 202,000 images, with 215 possible style traits.

\textbf{OpenViDial.} OpenViDial\footnote{\url{https://github.com/ShannonAI/OpenViDial}} \cite{meng2020openvidial} is an open-domain dialogue dataset with image contexts extracted from movies and TV series. There are a total of 1.1 million dialogue turns and 1.1 million visual contexts stored in the images.

\textbf{DodecaDialogue.} DodecaDialogue\footnote{\url{http://parl.ai/projects/dodecadialogue}} \cite{shuster2020dialogue} is a multi-task conversational benchmark, including 12 kinds of sub-datasets. Authors hope that such a multi-task training approach enables dialogue agents to ask questions, answer questions, utilize knowledge resources, discuss topics and contexts, and perceive and talk about images.

\subsubsection{Video-grounded Dialogue Datasets.}
As a relatively emerging research direction, there are only two video-based dialogue datasets to support the researches of video-based dialogue systems.

\textbf{AVSD.} Alamri \textit{et al.} \cite{alamri2019audio} firstly introduce the task of audio visual scene-aware dialog (AVSD), which requires dialogue agents to generate complete and natural responses to questions about the given video scenes. AVSD\footnote{\url{https://video-dialog.com}} contains 11,000 multi-turn conversations that discuss the content of videos about human daily activities from Charades dataset \cite{sigurdsson2016hollywood}.

\textbf{DVD.} DVD\footnote{\url{https://github.com/facebookresearch/DVDialogues}} \cite{le2021dvd} is a diagnostic video-grounded dialogue dataset, which contains detailed annotations for different types of reasoning over the spatio-temporal space of videos. The overall dataset consists of more than 100$k$ dialogues and 1 million question-answer pairs grounded with 11$k$ CATER synthetic videos \cite{girdhar2019cater}. DVD requires dialogue agents to have complex reasoning abilities on video and dialogue medium, such as action recognition, object reference, and spatio-temporal reasoning, which is a extremely challenging benchmark.

\subsection{Evaluation Metrics Towards VAD}
After training neural dialogue models, the another essential issue faced by researchers is how to precisely evaluate the performance of models and compare performances over multiple methods. Only with reasonable evaluation metrics, researchers can accurately measure the superiority or inferiority of designed models, to promote the progress of research communities. The quality of open domain dialogues with very high randomness and variability, are extremely difficult to be evaluated accurately. Up to now, there is no unified theory on how to effectively evaluate dialogue systems \cite{van2019best}, and some recognized metrics are adopted into the evaluation of VAD.

\subsubsection{Evaluation Metrics for Discriminative Models.} For discriminative decoder in VAD, the responses are evaluated in a retrieval setup. Specifically, at test time, every question is coupled with a list of candidate answers (e.g., 100 possible answers), and dialogue models need to sort candidate answers according to probabilities. The standard retrieval metrics, including Recall$@k$, Mean Reciprocal Rank (MRR) and Mean Rank, are usually used for evaluating the quality of discriminative dialogue models. Recall$@k$ represents the existence of the ground-truth response in top-$k$ ranked responses. MRR values placing ground-truth responses in higher ranks more heavily, and Mean Rank is sensitive to overall tendencies to rank ground-truth higher. The lower of MRR or the higher of Mean Rank indicate higher performance of dialogue systems.

\subsubsection{Evaluation Metrics for Generative Models.} For generative decoder in VAD, the response sequence to the given question are generated word by word, whose quality is evaluated by calculating the degree of coincidence with the ground-truth response. Some typical evaluation metrics are introduced as follows.

\textbf{BLEU.} BLEU \cite{papineni2002bleu} is the harmonic mean of $n$-gram word overlaps between generated responses and ground-truth reference responses, where $n\in \{1,2,3,4\}$. BLEU is the most intuitive way to calculate similarities between machine generated and target responses, but it also has many limitations, such as the tendency to assign higher scores to high frequency n-grams and shorter sentences \cite{doddington2002automatic}.

\textbf{METEOR.} METEOR \cite{lavie2007meteor} measures the precision and recall of unigrams between generated responses and ground-truth responses. To achieve more robust word matching, METEOR utilizes the fuzzy matching based on the stem analysis and WordNet synonym, to calculate the matching degree with multiple reference ground-truths.

\textbf{ROUGE-L.} Rather than calculating word-level overlaps, ROUGE-L \cite{lin2003automatic} proposes to evaluate the generated responses based on the length of the longest common subsequences (LCS) with the target responses. ROUGE-L scores are calculated from the F-measure of the maximum precision and recall of reference texts, where the accuracy and recall scores are obtained by dividing the length of LCS by the sequence length of generated responses and ground-truth.

\textbf{CIDEr.} CIDEr \cite{vedantam2015cider} considers semantic similarity in evaluating the quality of generated responses. On the level of $n$-grams, the average cosine similarity between generated responses and ground-truths is calculated to get CIDEr scores, where the importance of individual n-grams is determined by the Term Frequency Inverse Document Frequency (TF-IDF) measure.

\section{Open issues and future trends}
\label{chap:open}
Though researchers have made considerable efforts to address above challenges in VAD, there are still open issues to be resolved. In this section, we present some open issues and future development trends for VAD to promote the advancement of the research community.

%认知驱动的视频理解
%\subsection{Human Cognitive Science-based Visual Context Understanding}
\subsection{The Cognitive Mechanisms of Human-machine Dialogue Under Cross-modal Dialogue Context}
Existing studies adopt deep neural networks for visual and dialogue context feature extraction and understanding, which achieve certain appealing results. %However, the extracted visual or linguistic features still remain relatively superficial factual features, lacking deeper semantic understanding for visual scenes. 
However, these works focus only on the design of complex neural models and lack an intrinsic analysis of the cognitive mechanisms in human-machine dialogue interaction.
VAD expects dialogue agents to understand visual scenes like humans do to generate reasonable conversational responses. Therefore, it is necessary to research how to perceive and understand visual context for increasing the willingness and immersion of users during human-machine conversations from the cognitive mechanism level.
The research on people's cognitive mechanism has promising guidance for the visual information acquisition and understanding \cite{marr2010vision, yang2014cognitive} to obtain more comprehensive visual information. For example, the Dual-coding theory \cite{paivio2013imagery} indicates that our brain encodes information in two ways: \textit{visual imagery} and \textit{textual associations}. When referring to a concept, our brain either retrieves images or words, or both simultaneously. The ability to encode a concept in two different ways enhances our memory and understanding. Inspired by this cognitive theory, Jiang \textit{et al.} \cite{jiang2020dualvd} and Yu \textit{et al.} \cite{yu2020learning} propose to adaptively select question-relevant visual and context information in a hierarchical mode from multiple perceptual views and semantic levels. Moreover, Gatt \textit{et al.} \cite{gatt2013we} point out that human to be overspecific and prefer properties irrespectively when referring to objects. Based on this, Testoni \textit{et al.} \cite{testoni2021looking} propose the Confirm-it model to generate questions driven by the agent's confirmation bias for human-like dialogue generation.
A worthwhile future research point is to explore the cognitive mechanisms of human-machine dialogue under cross-modal dialogue context to provide sufficient priori knowledge for data-driven deep models, thus understanding visual context in a more efficient and informative way and generating more anthropomorphic responses.

%知识驱动的多模态对话系统
\subsection{Knowledge-enhanced Cross-modal Semantic Interaction}
In actual human-to-human conversations, we tend to subconsciously combine our mastered knowledge (e.g., commonsense knowledge and information about specific objects/events) to deeply understand dialogue context and the surrounding visual scenes \cite{huang2020challenges, guo2021conditional}. Knowledge is essential for acquiring deep semantic information and understanding the real world. Considerable research efforts have been made to combine external knowledge bases (e.g., ConceptNet \cite{speer2017conceptnet}, Wikipedia) to enhance the language understanding and generation in text-based dialogue systems \cite{kim2019sequential, ma2020survey, wang2021towards}. Up to now, very few works consider to fuse external knowledge in VAD to enhance the perception and understanding capabilities of visual and language context. For example, Liao \textit{et al.} \cite{liao2018knowledge} leverage fashion domain knowledge to capture the fine-grained semantics in images and generate informative responses. 
Furthermore, besides leveraging factual and encyclopedic knowledge about visual objects to enhance visual context understanding, the implicit logical semantics beyond the visual context are also should be considered. For instance, for videos containing humans, changes in sentiment or emotion of people are a key factor for visual context understanding \cite{soleymani2017survey, li2019survey}. Huber \textit{et al.} \cite{huber2018emotional} attempt to consider visual sentiment and facial expression of humans in visual scenes to generate meaningful and rich conversational interactions. 
To build engaging and harmonious VAD, it is indispensable to ground the concepts, entities and relations in textual dialogue history and visual context with external commonsense knowledge, real-world facts, and even human sentiment knowledge. With external knowledge as background information, VAD can comprehensively understand visual context, correlate dialogue context with visual semantics, and generate informative and engaging dialogue responses.

\subsection{Dialogue Attention Guided Adaptive Visual Context Understanding}
In the long-term human-human conversations, humans have the potential to quickly perceive a scene by selectively attending to concerned parts (e.g., specific objects or regions) and change conversational topics to fully explore and understand surroundings \cite{rensink2000dynamic, land2001ways}. This human attention helps to look critical visual context and reduce problem complexity, since the fixations can center on regions of interest and ignore background clutter. Inspired by this, some researches explore to use human attention regions in visual scenes to guide the machine's understanding process. For instance, Das \textit{et al.} \cite{das2017human} collect human attention maps of where humans choose to look in images to answer questions, which are used for explicitly supervising the question answer model to look at the same regions as humans. Capturing human attention is quite essential for fast and accurate understanding of visual scenes. Due to the constantly changing nature of human conversation topics, dialogue agents need to quickly understand user interaction intents and adaptively switch attention towards visual objects, regions or humans, to capture critical visual information for reasonable response generation.

%少样本 迁移学习

\subsection{Cross-modal Semantic Interaction with Limited Training Samples}
The training of neural dialogue models require massive dialogue samples to learn the human language expression with great variability. While for VAD, due to the hug semantic gaps between visual and language feature spaces, more massive amounts of video-grounded dialogue examples (i.e., dialogue turns with corresponding discussed images or videos) are needed to learn cross-modal semantic interactions for reasonable response generation, which are rarely available in any real-world platforms. 
Existing VAD datasets \cite{lin2014microsoft, alamri2019audio} are usually obtained by manual annotation on crowd-sourced platforms (e.g., Amazon Mechanical Turk (AMT)) and are very expensive and poorly scalable. To address the problem of unavailability of large-scale data, researchers consider few-shot or zero-shot learning in dialogue systems. For example, Li \textit{et al.} \cite{li2020zero} explore knowledge-enhanced dialogue systems without paired context-knowledge-response data and Song \textit{et al.} \cite{song2021bob} learn a consistent persona-based dialogue model from limited personalized dialogues. Zhao \textit{et al.} \cite{zhao2018zero} introduce a zero-shot dialogue system which can generalize to a new dialogue domain without the available training dialogues.
Through transfer knowledge from other domains or pre-trained large-scale models, dialogue systems have the potential to achieve better performance without sufficient training data. Integrating few-shot and zero-shot training paradigm for VAD is a promising research direction.

\subsection{Proactive Dialogue Guidance in VAD}
In human-human conversations, both parties usually chat actively by constantly and alternately looking for new topics, to make the conversation process highly participatory and interactive. Most of the existing dialogue systems focus on how to generate reasonable responses based on inputs, and this passive question-to-answer interaction mode will seriously reduce users' desire to communicate with dialogue agents \cite{li2016stalematebreaker}. 
The truly engaging and harmonious dialogue system should have the ability to proactively initiate dialogues with users, introduce new topics to attract users to communicate, and steer the dialogue process to achieve effective information sharing and cognitive guidance \cite{li2017dailydialog}. 
Proactive dialogue guidance can make a huge impact in many applications. For example, in the psychological de-escalation task, we expect VAD to understand users' psychological state progressively and relieve their psychological stress through proactive dialogue. Also through active information expression, VAD can realize the dissemination and guidance of online public opinion and change audiences' perceptions.
Some exploratory works attempt to empower dialogue systems with the ability to engage in proactive dialogue. For example, Liu \textit{et al.} \cite{liu2020speaker} propose a Initiative-Imitate model to force the dialogue agent to selectively act as a speaker or listener, thus achieving the flexible transition between proactive topic raising and passive interaction during conversation.
VAD should demonstrate the capability of cognitive guidance through proactive dialogue under cross-modal dialogue context to achieve target guidance goals, such as affective guidance, opinion guidance, etc.

%隐私安全
\subsection{Security of Privacy in VAD}
Through visual scenes perception, VAD brings more enjoyable human-machine interaction experiences for users. However, even for personal purposes, the visual perception process (e.g, taking photos or record videos of surroundings with mobile devices) creates a significant threat to the privacy of individuals in the footage. People are very sensitive to exposing private information about themselves to other people's images or videos, such as faces, license plates and home addresses. Therefore, the social concerns and ongoing debates about visual-based human-machine interaction systems are gradually emerging \cite{padilla2015visual, zhang2022visual}. Researchers have explored some privacy-protected video analytic methods to removing private information from videos \cite{yu2018pinto, wang2018enabling}. How to protect user privacy in VAD is a promising future research direction. Due to the localized nature of human conversational attention, which usually focuses only on specific visual regions, dividing frames into subimage blocks and pixelation is a way to protect visual privacy. 

\section{Conclusion}
\label{chap:conclusion}
We have made a systematic survey of visual-context augmented dialogue system (VAD). In this paper, we first introduce the concept and give the formalized definition of VAD. Based on the generic system architecture for VAD, we analyze the key challenges and comprehensively investigate representative works. The available datasets and evaluate metrics are also summarized. Through there are significant research progress in VAD, the researches are still in the early stages and extensive open issues and promising research trends need to be explored, such as the cognitive mechanisms of human-machine dialogue under cross-modal dialogue context, and knowledge-enhanced cross-modal semantic interaction.

\begin{acks}
This work was partially supported by the National Science Fund for Distinguished Young Scholars (62025205), and the National Natural Science Foundation of China (No. 62032020, 61960206008, 61725205).
\end{acks}

\bibliographystyle{ACM-Reference-Format}
\bibliography{VisualDial}


\end{document}